\documentclass[sigconf]{acmart}
\AtBeginDocument{%
  }

\setcopyright{acmlicensed}
\copyrightyear{2025}
\acmYear{2025}
\acmDOI{10.1145/3746027.3755362}

\copyrightyear{2025}
\acmYear{2025}
\setcopyright{cc}
\setcctype{by-nc}
\acmConference[MM '25]{Proceedings of the 33rd ACM International Conference on Multimedia}{October 27--31, 2025}{Dublin, Ireland}
\acmBooktitle{Proceedings of the 33rd ACM International Conference on Multimedia (MM '25), October 27--31, 2025, Dublin, Ireland}\acmDOI{10.1145/3746027.3755362}
\acmISBN{979-8-4007-2035-2/2025/10}
\usepackage{multirow}
\usepackage{colortbl}
\usepackage{adjustbox}

\usepackage{xcolor}
\definecolor{newcolor}{RGB}{0,47,167}

\usepackage{color}

\begin{document}

\title{Uni-DocDiff: A Unified Document Restoration Model Based on Diffusion}

\author{Fangmin Zhao}
\email{zhaofangmin@iie.ac.cn}
\orcid{0009-0000-1873-2603}
\affiliation{%
  \institution{Institute of Information Engineering, Chinese Academy of Sciences}
  \institution{School of Cyber Security, University of Chinese Academy of Sciences}
  \city{Beijing}
  \country{China}
}

\author{Weichao Zeng}
\email{weichaozeng22@gmail.com}
\orcid{0009-0000-4233-1201}
\affiliation{%
  \institution{Institute of Information Engineering, Chinese Academy of Sciences}
  \city{Beijing}
  \country{China}
}

\author{Zhenhang Li}
\email{zli74@binghamton.edu}
\orcid{0009-0002-6367-6692}
\affiliation{%
  \institution{Binghamton University}
  \city{Binghamton}
  \country{US}
}

\author{Dongbao Yang}
\email{yangdongbao@iie.ac.cn}
\orcid{0000-0001-8628-411X}
\affiliation{%
  \institution{Institute of Information Engineering, Chinese Academy of Sciences}
  \city{Beijing}
  \country{China}
}

\author{Binbin Li}
\authornote{Binbin Li and Yu Zhou are the corresponding authors.}
\email{libinbin@iie.ac.cn}
\orcid{0009-0002-4886-1267}
\affiliation{%
  \institution{Institute of Information Engineering, Chinese Academy of Sciences}
  \city{Beijing}
  \country{China}
}

\author{Xiaojun Bi}
\email{bixiaojun@hrbeu.edu.cn}
\orcid{0000-0002-5382-1000}
\affiliation{%
  \institution{Minzu University of China}
  \city{Beijing}
  \country{China}
}

\author{Yu Zhou}
\authornotemark[1]
\email{yzhou@nankai.edu.cn}
\orcid{0000-0003-4188-9953}
\affiliation{%
  \institution{VCIP \& TMCC \& DISSec, College of Computer Science, Nankai University}
  \city{Tianjin}
  \country{China}
}
\renewcommand{\shortauthors}{Zhao et al.}

\begin{abstract}
Removing various degradations from damaged documents greatly benefits digitization, downstream document analysis, and readability. Previous methods often treat each restoration task independently with dedicated models, leading to a cumbersome and highly complex document processing system. Although recent studies attempt to unify multiple tasks, they often suffer from limited scalability due to handcrafted prompts and heavy preprocessing, and fail to fully exploit inter-task synergy within a shared architecture. To address the aforementioned challenges, we propose Uni-DocDiff, a \textbf{Uni}fied and highly scalable \textbf{Doc}ument restoration model based on \textbf{Dif}fusion. Uni-DocDiff develops a learnable task prompt design, ensuring exceptional scalability across diverse tasks. To further enhance its multi-task capabilities and address potential task interference, we devise a novel \textbf{Prior \textbf{P}ool}, a simple yet comprehensive mechanism that combines both local high-frequency features and global low-frequency features. Additionally, we design the \textbf{Prior \textbf{F}usion \textbf{M}odule (PFM)}, which enables the model to adaptively select the most relevant prior information for each specific task. Extensive experiments show that the versatile Uni-DocDiff achieves performance comparable or even superior performance compared with task-specific expert models, and simultaneously holds the task scalability for seamless adaptation to new tasks.
\end{abstract}




\maketitle

\section{Introduction}

\begin{figure*}[ht]
\centering
\includegraphics[width=1\textwidth]{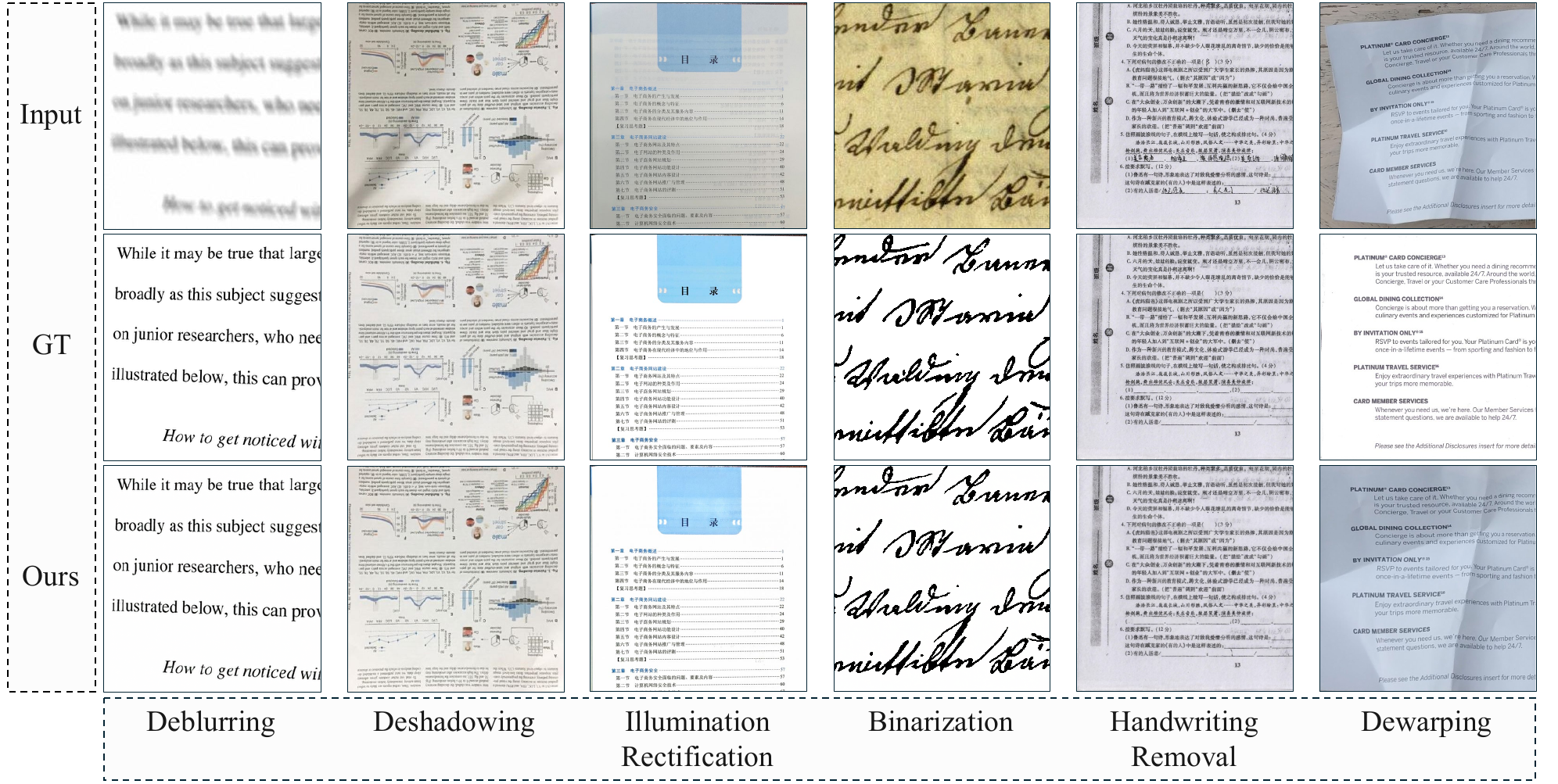}
\caption{Examples of our proposed Uni-DocDiff on tasks including deblurring, deshadowing, illumination recitification, binarization, handwriting removal and dewarping. It is capable of effectively handling multiple tasks..}
\label{overview}
\end{figure*}

Document images often suffer from various degradations, including deformations, handwritten traces, uneven illumination, and other capturing issues. This not only significantly impacts the users' reading experience but also hinders the performance of downstream document analysis \cite{shen2025ldp,shen_ijcai,zhang2025class}, recognition \cite{zheng2024cdistnet,qiao2020seed,yang2025ipad} and spotting \cite{lyu2025arbitrary,wang2022tpsnet,li2025beyond} systems, as most of these systems are trained on clean, scanned documents. Consequently, a multitude of approaches have been proposed to tackle individual degradations, such as dewarping \cite{ma2018docunet,xie2021document,feng2022geometric,verhoeven2023uvdoc,shu2025visual}, deblurring \cite{mamidibathula2019svdocnet,mei2019deepdeblur}, deshadowing \cite{lin2020bedsr,zhang2023document,li2023high}, illumination rectification \cite{wang2022udoc,zhang2023appearance,hertlein2023template}, binarization \cite{biswas2023docbinformer,yang2024gdb}, and handwritting removal \cite{wang2022chenet,huang2023ensexam,li2024scene}. However, these single-task models exhibit insufficient generalizability when faced with real-world multi-degradation documents.

The recently proposed DocRes \cite{zhang2024docres} marks a significant advancement in the development of a unified method for document restoration, demonstrating impressive performance across various tasks. Nonetheless, its reliance on meticulously crafted prompts constrains its scalability to adapt to new types of image degradations without extensive retraining. Additionally, the interactions and sharing mechanisms between tasks have not been explored.

\begin{figure}[tb]
\centering
\includegraphics[width=0.5\textwidth]{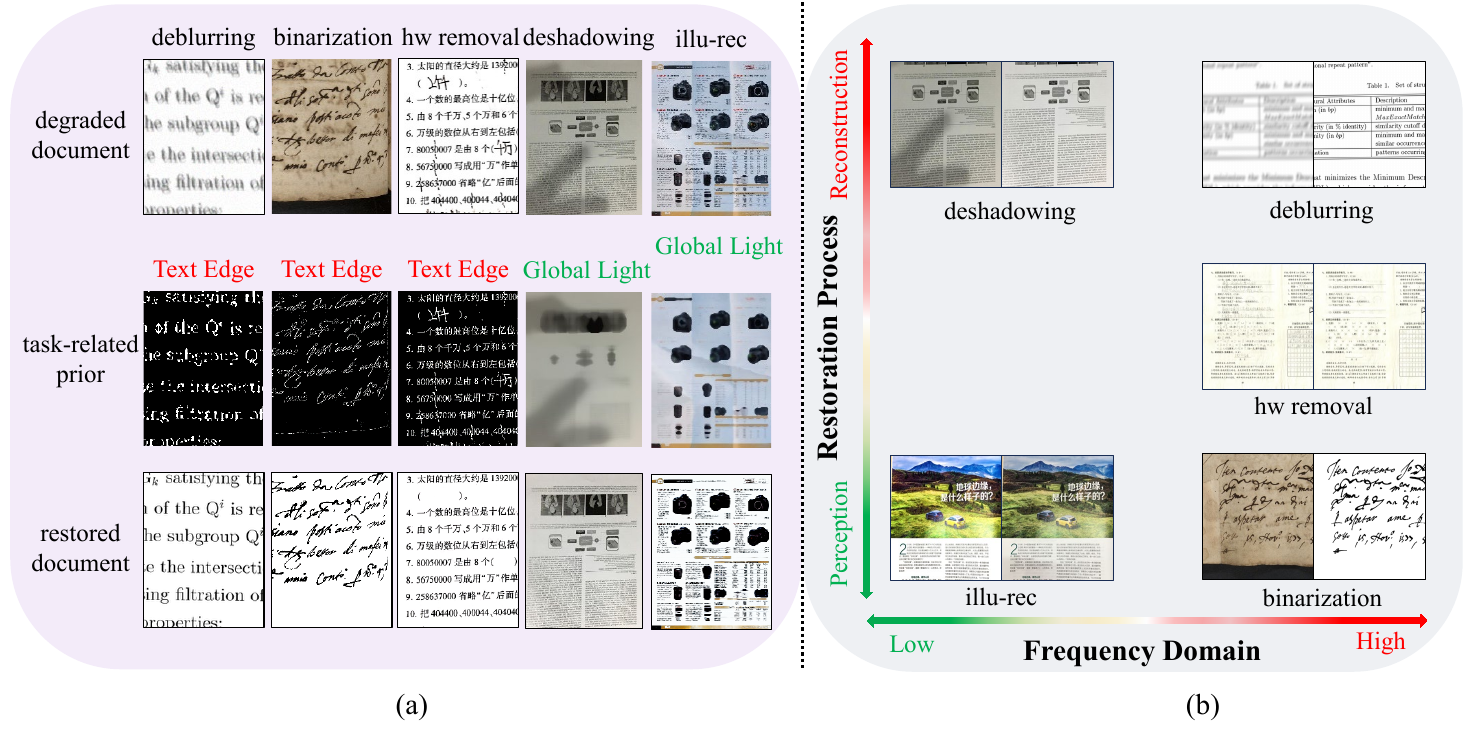}
\caption{Each task has different focuses on feature. Zoom in to view. (a) Visualization of the features focused upon by different tasks. (b) Different tasks focus on respective frequency-domain features at different restoration stages.}
\label{prior}
\end{figure}

In fact, different tasks share similar or opposite feature focus during the restoration process. As shown in Fig. \ref{prior} (a), deblurring, binarization, and handwriting removal focus on high-frequency features, whereas deshadowing and illumination rectification emphasize low-frequency features. Specifically, as shown in Fig. \ref{prior} (b), deblurring requires constructing high-frequency details such as the edges of text; binarization perceives high-frequency features to determine the strokes of the text; and handwriting removal uses high-frequency features to simultaneously extract printed text and handwritten script, then further distinguishes between them, thereby removing the handwritten script while preserving the printed text. On the other hand, deshadowing constructs global low-frequency content such as background illumination. In contrast, illumination rectification perceives low-frequency feature and eliminates it. As a result, integrating these tasks into a single model can lead to interference during training, especially due to differences in gradient updates. This can hinder model stability and convergence, ultimately leading to performance that falls short of specialized, task-specific models. This presents a significant challenge when aiming for unified restoration solutions.

To address this issue, we propose a simple yet effective solution: a Prior Pool that stores task-specific prior features, along with a Prior Fusion Module (PFM) that enables the model to adaptively select and emphasize the most relevant features from the Prior Pool. By integrating these selected prior features with the shared document features across multiple tasks, our approach minimizes task interference and enhances overall performance.

\begin{figure}[tb]
\centering
\includegraphics[width=0.45\textwidth]{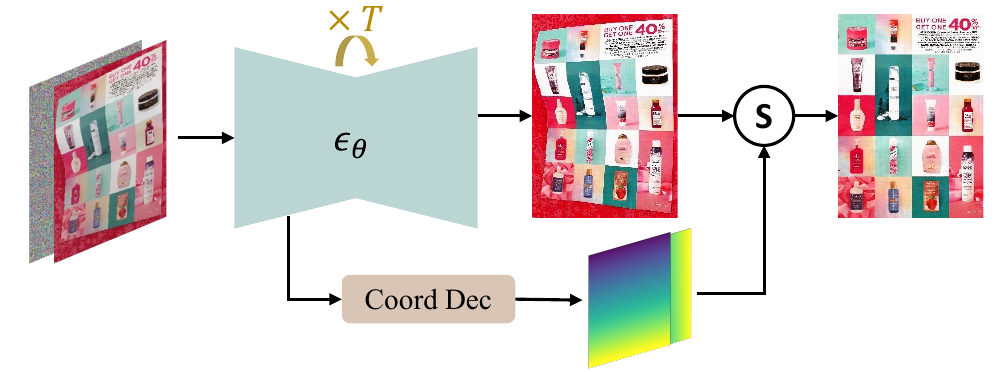}
\caption{Schematic illustration of our \textit{Uni-DocDiff}.}
\label{dualstream}
\end{figure}

Due to the "regression to the mean" inherent in regression methods \cite{zhang2024docres}, restored text edges often appear blurred or damaged. Considering the ability of diffusion models \cite{ho2020denoising,song2020denoising,zeng2024textctrl,li2024first} to generate high-quality images and high-frequency details, we propose a unified document restoration model based on diffusion. However, dewarping differs from tasks for pixel restoration. For dewarping, it is unnecessary to reconstruct high-frequency details iteratively. A simple decoder can efficiently achieve the desired results without wasting resources. Therefore, we decouple the coordinate space from the pixel space and propose a dual-stream Uni-DocDiff, as shown in Fig. \ref{dualstream}, which learns the coexisting components, reduces interference between tasks, and lowers the learning cost.

Our key contributions are summarized as follows:

\begin{itemize}
\item 
We propose a \textbf{Uni}fied \textbf{Doc}ument image restoration model based on \textbf{Diff}usion, called Uni-DocDiff. To the best of our knowledge, this is the first unified document restoration model to apply diffusion models in document processing.
\item We design a simple yet comprehensive Prior Pool that includes low-level local texture information and high-level global semantic information and introduce a \textbf{P}rior \textbf{F}usion \textbf{M}odule to eliminate task interference.
\item 
Extensive experiments demonstrate that Uni-Docdiff achieves superior performance compared to task-specific restoration models and exhibits excellent task scalability, a capability that existing methods have not achieved.
\end{itemize}

\section{Related Work}

\subsection{Document Image Restoration}

In real-world scenarios, document images frequently suffer from various degradation. Numerous methods focusing on individual degradation types.
Dewarping methods\cite{zhang2023polar,liu2023rethinking,verhoeven2023uvdoc,feng2022geometric,li2023layout,li2023foreground,ma2018docunet,feng2023deep} focus on addressing geometric distortion predicting the coordinate mapping between distorted documents and flat documents. For deshadowing, BEDSR-Net \cite{lin2020bedsr} predicts global background and encodes shadow position as an attention map. FSENet \cite{li2023high} utilizes low-frequency details and high-frequency edges to handle high-resolution document shadows. For illumination rectification, GCDRNet \cite{zhang2023appearance} sequentially learns global content and recovers details. UDoc-GAN \cite{wang2022udoc} improves upon CycleGAN \cite{cyclegan} to train with unpaired data. SVDocNet \cite{mamidibathula2019svdocnet} combines CNN and RNN for deblurring, while DocDiff \cite{yang2023docdiff} utilizes diffusion to generate high-frequency residual details. EnsExam \cite{huang2023ensexam} introduces a soft stroke mask and MSFF-Net \cite{li2024scene} extracts multi-scale feature to erase handwritten text. For binarization, GDB \cite{yang2024gdb} extracts high-frequency prior, providing additional prior. DocBinFormer\cite{biswas2023docbinformer} utilizes a two-level Transformer to extract local and global features.

Some methods \cite{li2019document,feng2021doctr,das2019dewarpnet} sequentially address dewarping and illumination rectification. Tang et al. \cite{tang2024efficient} adopt a dual-stream network to handle geometry and lighting separately, while DocNLC \cite{wang2024docnlc} uses contrastive learning for unified restoration. Though multi-task, these approaches remain limited in task diversity and real-world generalization.

\begin{figure*}[ht]
\centering
\includegraphics[width=1\textwidth]{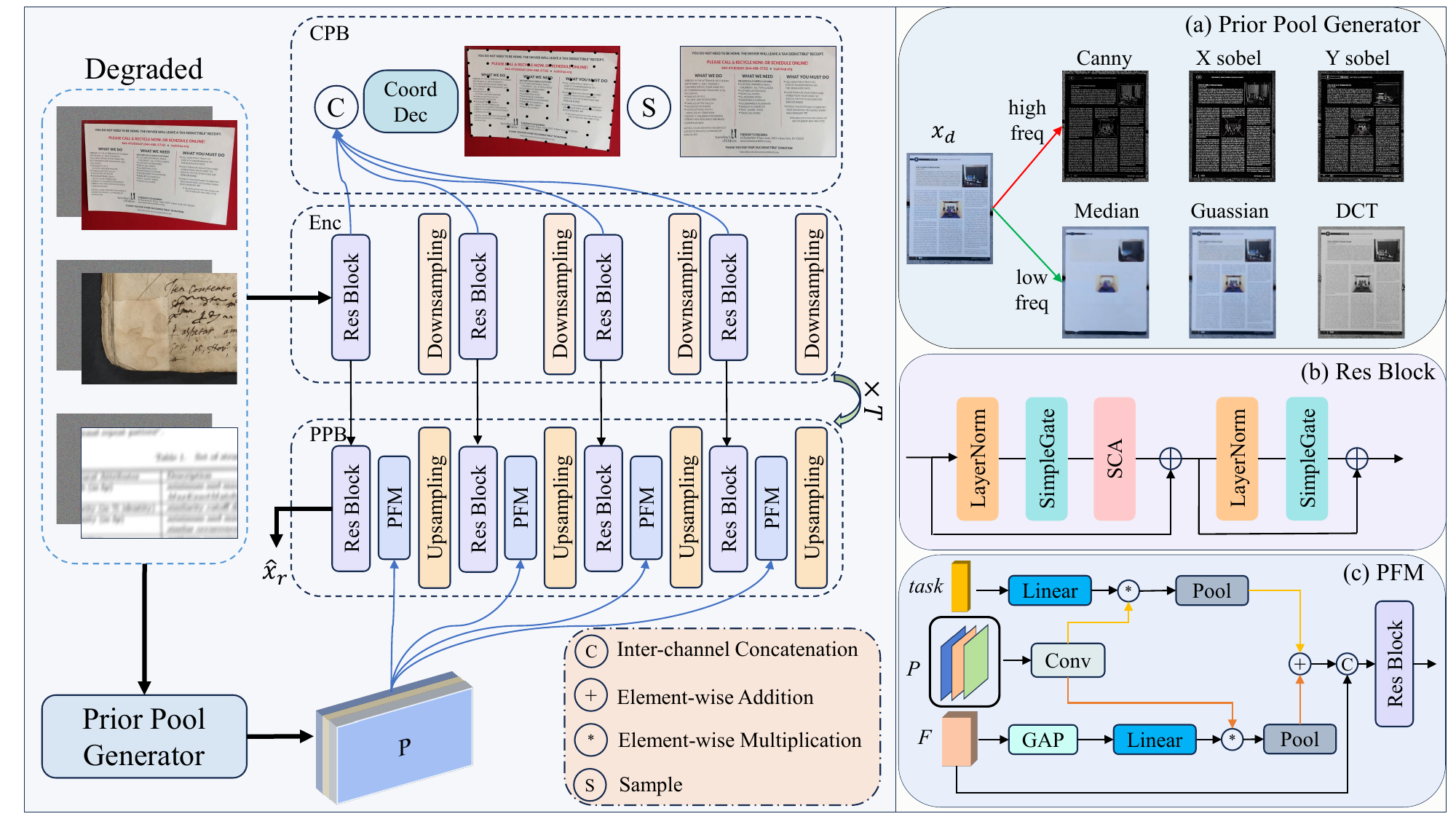}
\caption{An overview of the proposed Uni-DocDiff, which contains a diffusion-based pixel prediction branch and a light-weight coordinate prediction branch.}
\label{overview}
\end{figure*}

\subsection{All-in-one image restoration}


DocRes \cite{zhang2024docres} is the only notable work on unified document image restoration. It proposes DTSPrompt to distinguish tasks and incorporate priors, marking a successful step toward unified restoration. In contrast, unified models for natural scenes are more diverse, covering deraining, dehazing, deblurring, and beyond.

Blind image restoration \cite{lin2025diffbir,kulkarni2022unified,potlapalli2024promptir} is often applied in scenarios such as weather conditions removal tasks. However, for document images restoration, similar inputs might require different outputs, such as deshadowing and illumination rectification, which can lead to confusion between tasks, failing to accomplish specific restoration.
Task-aware unified image restoration often requires additional prompts to guide the model in handling different types of degradation, including learnable prompts, visual prompts, or text prompts. ProRes \cite{ma2023prores} uses learnable prompts, adding task-specific prompts to the input image before feeding it into the restoration network. PromptGIP\cite{liu2023unifying} employs a distorted-clean image pair as visual prompts, enabling the model to perceive the distortion type. During the training process, a random portion of patches is masked and then restored. Works such as SPIRE \cite{qi2025spire}, DiffPlugin \cite{liu2024diff}, and InstructIR \cite{conde2024high} utilize text prompt, encode user instructions to extract intent and feed it into the diffusion model to guide the generation process.

Despite the notable progress of existing unified image restoration methods in natural scenes, their performance \cite{ma2023prores,potlapalli2024promptir} still lags behind that of task-specific models. Moreover, unified restoration for document scenes still lacks exploration. Therefore, we proposes Uni-DocDiff to fill this research gap.

\section{Methodology}

\subsection{Overview}
Given a specified degradation
type, Uni-DocDiff aims to transform the degraded image $x_d \in \mathbb{R}^{H \times W \times 3}$ into a restored image $x_r \in \mathbb{R}^{H \times W \times 3}$.  Fig. \ref{overview} shows the overall architecture of Uni-DocDiff, a dual-stream model comprising a Coordinate Prediction Branch (CPB) for dewarping and a Pixel Prediction Branch (PPB) for other restoration tasks. The CPB predicts a sparse \textit{backward map} $f_{bm}$, representing coordinate positions of reference points in the flattened image. And the PPB performs task-specific pixel-wise restoration.

To enhance performance across diverse restoration tasks, Uni-DocDiff incorporates a designed Prior Pool. Through the learnable Prior Fusion Module (PFM), the model adaptively assigns task-relevant and content-based prior weights, effectively handling various restoration challenges while mitigating inter-task interference. The following subsections detail: (1) the PPB branch, including the model architecture, Prior Pool design, and PFM structure; (2) the CPB branch; and (3) the training pipeline.

\subsection{Pixel Prediction Branch}

PPB uses a diffusion model as its backbone and constructs Prior Pool and PFM to provide priors and differentiate tasks, respectively. We will introduce these components in detail below.

\textit{Denoiser $\mathcal{F}$.} Similar to mainstream diffusion models \cite{rombach2022high}, our denoiser $\mathcal{F}$ is a image-conditional diffusion model. It includes multiple residual convolutional blocks \cite{chen2022simple} and employs skip connection operations to increase the retention of information. To save computational costs, we only incorporate the transformer structure in the middle block.
During the forward process, we randomly sample a time step $t \sim Uniform(0,T_{max})$ to noise the clean GT image $x_0$ and obtain the noised image $x_t$, as follows:

\begin{equation}
    x_t = \sqrt{\overline{\alpha}_t}x_0+\sqrt{1-\overline{\alpha}_t}\epsilon_t
\end{equation}
where $\alpha_t$ is a hyperparameter, $\alpha_0=1$, and $\overline{\alpha}_t = \prod_{i=0}^t \alpha_i.$ 

Then we concatenate the noised image $x_t \sim \mathbb{R}^{3 \times H \times W}$ and the degraded image $x_d \sim \mathbb{R}^{3 \times H \times W}$ in channel dimension as a 6-channel input. As detailed in DocDiff\cite{yang2023docdiff}, for the target of model prediction, while predicting noise $\epsilon$ and predicting $x_0$ are equivalent in the case of unconditional generation, predicting $x_0$ tends to reduce diversity but can enhance generation quality for conditional generation. Therefore, we opt to predict $x_0$. Consequently, the inference process we employ can be formulated as follows:
\begin{equation}
    \Hat{x}_0 = \mathcal{F}(x_t; x_d, task, P)
\end{equation}

Here, $task$ refers to the current task being processed, and $P$ denotes the Prior Pool generated based on the distorted image $x_d$. And the reverse denoising process is:

\begin{equation}
    x_{t-1} = \sqrt{\overline{\alpha}_{t-1}}\hat{x}_0+\sqrt{1-\overline{\alpha}_{t-1}} \cdot \frac{x_t-\sqrt{\overline{\alpha}_t}\hat{x}_0}{\sqrt{1-\overline{\alpha}_t}}
\end{equation}

\textit{Prior Pool.} In order to alleviate the mutual interference between tasks and ensure the restoration performance of all-in-one models, we propose a simple yet comprehensive Prior Pool Generator, as shown in the Fig. \ref{overview}, these priors $P$ consist of a series of features that have the same resolution as the original image. Specifically, we utilize simple, convenient, and time-efficient traditional image feature extraction operators to obtain the high-frequency and low-frequency features of the distorted images. For the extraction of high-frequency features, we employ the Sobel operator to compute the first-order derivative images in both horizontal and vertical directions, while the Canny operator is utilized to capture edge features from the images. It is well-known that Sobel-based images are highly sensitive to noise, whereas Canny-based images are prone to information loss due to improper threshold selection. By combining these two approaches, the model can achieve a more comprehensive understanding and reconstruction of high-frequency features, including document edges and text boundaries, thereby enhancing its overall performance in feature perception. For handling low-frequency features, median filtering is applied to remove salt-and-pepper noise and sparse text pixels from the images. We also use Gaussian filtering to generate a smoothly continuous background illumination map. Moreover, by employing the Discrete Cosine Transform (DCT), crucial low-frequency components and local texture edge features are retained. This methodology enables the model to estimate and reconstruct background illumination with greater accuracy, thereby enhancing its capacity to faithfully represent the image's critical attributes.

\textit{PFM.} To enable the model to fully leverage the information contained in the Prior Pool $P$, and to adaptively select relevant features based on the task characteristics and the reconstruction phase, we propose a novel Prior Fusion Module. PFM is strategically positioned between Res Block and the Upsampling Block, allowing the model to dynamically assess the importance of the Prior Pool according to the stage of reconstruction. Its primary structure is illustrated in the lower right of the Fig. \ref{overview}. Specifically, we leverage the image features $f^l \in \mathbb{R}^{h_l \times w_l \times c_l}$, $task$, and Prior Pool $P \in \mathbb{R}^{H \times W \times C}$ at each stage to derive task-specific reconstruction features $f_{recon}^l \in \mathbb{R}^{h_l \times w_l \times c_l}$, where $h_l, w_l, c_l$ refer to the height, width, and number of channels of the image features at the $l$-th stage of the reconstruction process and $l \in \{1,2,3,4\}$. Firstly, to enable the $P$ to adapt to the feature distributions at different stages of reconstruction, we first construct multiple layers of residual convolutional layers. This processing yields a refined prior $P^l = Conv(P), P^l \in \mathbb{R}^{h_l \times w_l \times c_l}$, which aligns with the feature distribution specific to each stage. Subsequently, a Global Average Pooling (GAP) operation followed by multiple linear layers is applied to process $f^l$, yielding content-specific weights $w_{content}^l \in \mathbb{R}^{c_l}$, and several linear layers are applied to process $task$ to obtain task-specific weights $w_{task}^l \in \mathbb{R}^{c_l}$:

\begin{equation}
    w_{task}^l = MLP(task), w_{content}^l = MLP(GAP(f^l))
\end{equation}

Then these two weights are individually multiplied with $P^l$ to obtain content-specific priors $P_{content}^l$ and task-specific priors $P_{task}^l$. These priors are then summed and concatenated with $f^l$ along the channel dimension, and then is fed into the subsequent Res Block to obtain the final reconstructed features $f_{recon}^l$:

\begin{equation}
    f_{recon}^l = ResBlock(Concat(f^l,P_{task}^l+P_{content}^l))
\end{equation}

Therefore, the localization of interactions between features $F$, tasks $T$, and the prior pool $P$ exclusively within the PFM architecture enhances task flexibility and scalability. When integrating novel tasks, the diffusion backbone can be frozen to preserve established knowledge, necessitating training solely for the PFM. This approach significantly reduces training complexity. Furthermore, the trainability of PFM enables selective modification of the Prior Pool according to specific characteristics of emerging tasks, thereby facilitating optimal performance and adaptability.

\subsection{Coordinate Prediction Branch}

\begin{figure}[tb]
\centering
\includegraphics[width=0.45\textwidth]{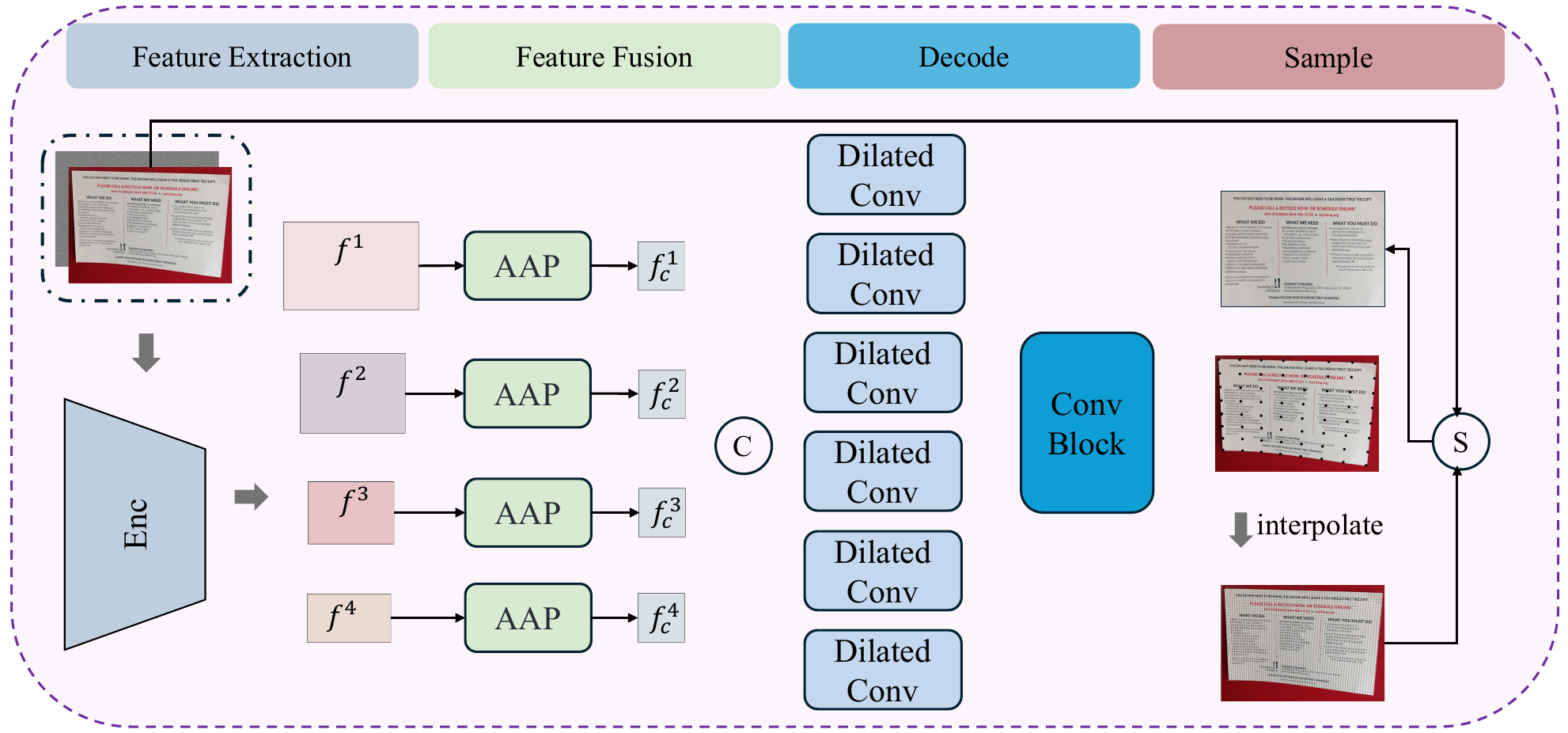} 
\caption{Structure Diagram of the CPB. }
\label{cpb}
\end{figure}

Since the output space for coordinate prediction does not require multiple iterations of the diffusion model to achieve high-frequency perception, and numerous studies \cite{baranchuk2021label} have demonstrated the excellent feature extraction capabilities of diffusion model encoders, we construct a dual-stream architecture. The specific structure is illustrated in the Fig. \ref{cpb}. 

Specifically, the image is processed through the Encoder to extract multi-scale features $f^l, l \in \{1, 2, 3, 4\}$. These features are then subjected to Adaptive Average Pooling (AAP) to obtain fixed-size feature maps $f_c^l$, facilitating subsequent multi-scale feature fusion and model localization. $f_c^l$ are concatenated along the channel dimension to form fused coordinate prediction features $f_c$. Inspired by the DDCP\cite{xie2021document}, we construct six dilated convolution blocks with different dilation rates to process $f_c$, with dilation rates of 1, 2, 5, [8, 3, 2], [12, 7, 4], and [18, 12, 6], respectively, obtaining degradation-aware features $f_d^i , i \in \{1, 2, 3, 4, 5, 6\}$ with varying receptive fields. 
$f_d^i$ is concatenated and a simple three-layer convolutional module is subsequently applied to produces the final sparse control point matrix (i.e., the backward map or $bm$. The $bm$ matrix is then upsampled to the original image resolution and used to sample from the distorted image, thereby generating a flattened document image. The procedure can be formulated as:

\begin{equation}
\begin{aligned}
f_c^l &= AAP (f^l) \\
f_c &= Concat(f_c^1, f_c^2, f_c^3, f_c^4) \\
f_d^i &= DilateConvBlock_i(f_c) \\
bm &= ConvBlock(Concat(f_d^1, f_d^2,f_d^3,f_d^4,f_d^5,f_d^6))
\end{aligned}
\end{equation}

Through this design, we achieve high-quality image pixel restoration as well as efficient and straightforward image geometry dewarping.

\begin{table*}[ht]
    \centering
    \setlength{\tabcolsep}{0.6mm}
    \caption{Quantitative comparison results between our Uni-DocDiff model and existing task-specific and previous unified state-of-the-art models on various tasks. The tasks, from top to bottom, are: deblurring, deshadowing, illumination rectification, dewarping, binarization, and handwriting removal. ``$\uparrow$'' and ``$\downarrow$'' signifies higher
and lower better respectively. The \textbf{best} results are represented in bold and the \underline{second-best} results are underlined. ``*'' indicates that we use the DocUNet dataset dewarped by the DocAligner model for evaluation. The same applies to the tables below.}
    \scriptsize
    \begin{tabular}{c|c|cccccccccccc|c}
        \toprule
        \multirow{2}{*}{{\begin{tabular}{c} Metrics \end{tabular}}} & \multirow{2}{*}{{\begin{tabular}{c} Datasets \end{tabular}}} & \multicolumn{1}{c}{DE-GAN\cite{souibgui2020gan}} & \multicolumn{1}{c}{DocDiff\cite{yang2023docdiff}} & \multicolumn{1}{c}{BGSNet\cite{zhang2023document}} & \multicolumn{1}{c}{DocShadow\cite{li2023high}} & \multicolumn{1}{c}{GCDRNet\cite{zhang2023appearance}} & \multicolumn{1}{c}{UDoc-GAN\cite{wang2022udoc}} & \multicolumn{1}{c}{UVDoc\cite{verhoeven2023uvdoc}} & \multicolumn{1}{c}{DocGeo\cite{feng2022geometric}} & \multicolumn{1}{c}{GDB\cite{yang2024gdb}} & \multicolumn{1}{c}{EnsExam\cite{huang2023ensexam}} & \multicolumn{1}{c}{CTRNet \cite{liu2022don}} & \multicolumn{1}{c}{DocRes\cite{zhang2024docres}} & \multicolumn{1}{c}{Ours}\\
        & & \multicolumn{1}{c}{\textit{TPAMI'20}} & \multicolumn{1}{c}{\textit{MM'23}} & \multicolumn{1}{c}{\textit{CVPR'23}} & \multicolumn{1}{c}{\textit{ICCV'23}} & \multicolumn{1}{c}{\textit{TAI'23}} & \multicolumn{1}{c}{\textit{MM'22}} & \multicolumn{1}{c}{\textit{SIGGRAPH'23}} & \multicolumn{1}{c}{\textit{ECCV'22}} & \multicolumn{1}{c}{\textit{PR'24}} & \multicolumn{1}{c}{\textit{ICDAR'23}} & \multicolumn{1}{c}{\textit{ECCV'22}} & \multicolumn{1}{c}{\textit{CVPR'24}} & \multicolumn{1}{c}{ }\\
        \midrule
        \multirow{2}{*}{{\begin{tabular}{c} SSIM $\uparrow$ \\ PSNR $\uparrow$ \end{tabular}}} & \multirow{2}{*}{TDD\cite{hradivs2015convolutional}} & 0.9226 & 0.9559 & - & - & - & - & - & - & - & - & - & \underline{0.9723} & \textbf{0.9824} \\
        & & 22.24 & 24.00  & - & - & - & - & - & - & - & - & - & \underline{27.35} & \textbf{28.77} \\
        \midrule
        \multirow{6}{*}{{\begin{tabular}{c} SSIM $\uparrow$ \\ PSNR $\uparrow$ \end{tabular}}} & \multirow{2}{*}{Jung's\cite{jung2018water}} & - & - & 0.9094 & 0.9005 & - & - & - & - & - & - & - & \underline{0.9089} & \textbf{0.9156} \\
        & & - & - & 17.34 & 21.05 & - & - & - & - & - & - & - & \underline{23.02} & \textbf{23.93} \\
        & \multirow{2}{*}{Kligler's\cite{kligler2018document}} & - & - & \textbf{0.9480} & 0.9088 & - & - & - & - & - & - & - & 0.9005 & \underline{0.9382} \\
        & & - & - & \textbf{29.17} & 25.12 & - & - & - & - & - & - & - & 27.14 & \underline{28.56} \\
        & \multirow{2}{*}{OSR\cite{kligler2018document}} & - & - & \underline{0.9487} & 0.9018 & - & - & - & - & - & - & - & 0.9394 & \textbf{0.9532} \\
        & & - & - & \textbf{22.31} & 18.25 & - & - & - & - & - & - & - & 19.17 & \underline{21.48} \\
        \midrule
        \multirow{4}{*}{{\begin{tabular}{c} SSIM $\uparrow$ \\ PSNR $\uparrow$ \end{tabular}}} & \multirow{2}{*}{DocUNet*\cite{ma2018docunet}} & - & - & - & - & \underline{0.7658} & 0.6833 & - & - & - & - & - & 0.7598 & \textbf{0.7682} \\
        & & - & - & - & - & 17.09 & 14.29 & - & - & - & - & - & \underline{17.60} & \textbf{18.22} \\
        & \multirow{2}{*}{RealDAE\cite{zhang2023appearance}} & - & - & - & - & \underline{0.9423} & 0.7558 & - & - & - & - & - & 0.9219 & \textbf{0.9485} \\
        & & - & - & - & - & 24.42 & 16.43 & - & - & - & - & - & \underline{24.65} & \textbf{24.97} \\
        \midrule
        \multirow{3}{*}{{\begin{tabular}{c} MSSSIM $\uparrow$ \\ LD $\downarrow$ \\ AD $\downarrow$ \end{tabular}}} & \multirow{3}{*}{DIR300\cite{feng2022geometric}} & - & - & - & - & - & - & 0.6216 & \underline{0.6380} & - & - & - & 0.6264 & \textbf{0.6573} \\
        & & - & - & - & - & - & - & 7.73 & \underline{6.40} & - & - & - & 6.83 & \textbf{5.30} \\
        & & - & - & - & - & - & - & \underline{0.218} & 0.242 & - & - & - & 0.241 & \textbf{0.203} \\
        \midrule
        \multirow{3}{*}{{\begin{tabular}{c} FM $\uparrow$ \\ pFM $\uparrow$ \\ PSNR $\uparrow$ \end{tabular}}} & \multirow{3}{*}{DIBCO'18\cite{pratikakis2018icfhr}} & - & 88.11 & - & - & - & - & - & - & \textbf{91.09} & - & - & 89.82 & \underline{90.32} \\
        & & - & 90.43 & - & - & - & - & - & - & \textbf{94.57} & - & - & \underline{94.33} & 93.84 \\
        & & - & 17.92 & - & - & - & - & - & - & \textbf{19.92} & - & - & 19.35 & \underline{19.76} \\
        \midrule
        \multirow{2}{*}{{\begin{tabular}{c} MSSSIM $\uparrow$ \\ PSNR $\uparrow$ \end{tabular}}} & \multirow{2}{*}{EnsExam\cite{huang2023ensexam}} &  - & - & - & - & - & - & - & - & - & 0.9659 & \underline{0.9671} & - & \textbf{0.9685} \\
        & & - & - & - & - & - & - & - & - & - & \underline{36.05} & 35.68 & - & \textbf{36.23} \\
        
        \hline
    \end{tabular}
    \label{sota_compare}
\end{table*}

\subsection{Training Pipeline}

Our training pipeline consists of two stages:

\textbf{Stage 1} primarily focuses on the training of the Encoder and PPB, thereby handling multiple restoration tasks, including binarization, deblurring, deshadowing, illumination rectification, and handwriting removal. We design loss functions based on the specific characteristics of each task, including the $L1$ loss between the predicted image and the reference image, as well as a frequency-aware loss. For deshadowing and illumination rectification, we design a low-frequency-aware loss $\mathcal{L}_{lowfreq}$. And for deblurring, binarization and handwriting removal, we design a high-frequency-aware loss $\mathcal{L}_{highfreq}$:
\begin{equation}
\begin{aligned}
    \mathcal{L}_{lowfreq} &= \mathbb{E} \left\| \phi_{\mathrm{L}}(\hat{x}_0) - \phi_{\mathrm{L}}(x_0) \right\| \\
    \mathcal{L}_{highfreq} &= \mathbb{E} \left\| \phi_{\mathrm{H}}(\hat{x}_0) - \phi_{\mathrm{H}}(x_0) \right\|
\end{aligned}
\end{equation}

where $\phi_{\mathrm{L}}$ is the low-pass filter and $\phi_{\mathrm{H}}$ is the high-pass filter. The overall loss for the two types of tasks can be expressed as $\mathcal{L}_{low}$ and $\mathcal{L}_{high}$:

\begin{equation}
\begin{aligned}
    \mathcal{L}_{low} &= \mathbb{E} \left\| \hat{x}_0 - x_0 \right\| + \beta_1\mathcal{L}_{lowfreq} \\
    \mathcal{L}_{high} &= \mathbb{E} \left\| \hat{x}_0 - x_0 \right\| + \beta_2\mathcal{L}_{highfreq}
\end{aligned}
\end{equation}

$\beta_1$ and $\beta_2$ are hyperparameters used to control the numerical magnitude of the loss functions for different tasks, ensuring they are close to each other.

\textbf{Stage 2} focuses on training the CPB. Due to the training in the first stage, the Encoder is equipment with the ability to extract features. Therefore, we freeze the parameters of the Encoder and utilize the dewarping dataset to train the CPB branch. The loss function $\mathcal{L}_c$ is:

\begin{equation}
    \mathcal{L}_c = \mathbb{E} \left\| bm - bm_{gt} \right\|
\end{equation}

Through the two-stage training, our model has achieved the capability to handle multiple restoration tasks while also possessing the ability to add new tasks at a low cost.

\section{Experiment}

\subsection{Datasets}

For \textit{Dewarping}, the Doc3D dataset \cite{das2019dewarpnet}, which contains 100,000 synthetic images, and UVDoc dataset \cite{verhoeven2023uvdoc}, which contains 20K synthetic images, are used for training. For testing, the DocUNet \cite{ma2018docunet} and DIR300 \cite{feng2022geometric} are utilized, comprising 130 and 300 real-world images, respectively.

For \textit{Deblurring}, following DocRes \cite{zhang2024docres}, we randomly sample 40K images for training from the Text Deblur Dataset (TDD) \cite{hradivs2015convolutional}, and 1.6K images are selected for evaluation.

For \textit{Deshadowing}, we use SD7K \cite{li2023high} and the training set from RDD \cite{zhang2023document} for deshadowing training. FSDSRD and the training sets of RDD contain 14,200 synthetic images and 4,916 real images, respectively. Jung’s dataset \cite{jung2018water}, Kligler’s dataset \cite{kligler2018document} and OSR \cite{wang2020local} are utilized for testing, with 87, 300, 237 images, respectively. 

For \textit{Illumination Rectification}, for training, we synthesize a dataset by combining illumination images from DocShade \cite{das2020intrinsic} with collected clean document images, and include 450 real images from the RealDAE training set. For testing, we utilize the RealDAE test set and the DocUNet model after dewarping by DocAligner \cite{zhang2023docaligner}.

For \textit{Binarization}, we collect the datasets from DIBCO(09-17) \cite{gatos2009icdar,pratikakis2010h,pratikakis2012icfhr,pratikakis2013icdar,ntirogiannis2014icfhr2014}, DIBCO'19\cite{pratikakis2017icdar2017}, the Noisy Office Dataset \cite{zhang2023all}, MSI \cite{hedjam2015icdar}, PHIBD \cite{nafchi2013efficient}, and Bickley Diary Dataset \cite{deng2010binarizationshop} as the training set, and use DIBCO2018 \cite{pratikakis2018icfhr} as the test set.

For \textit{Handwriting Removal}, the data used for training originates from the Handwritten Erasure Competition dataset and the EnsExam \cite{huang2023ensexam} training set, comprising a total of 26,895 images. For testing, we utilize the EnsExam test set, which consists of 115 images.

\begin{figure*}[ht]
\centering
\includegraphics[width=1\textwidth]{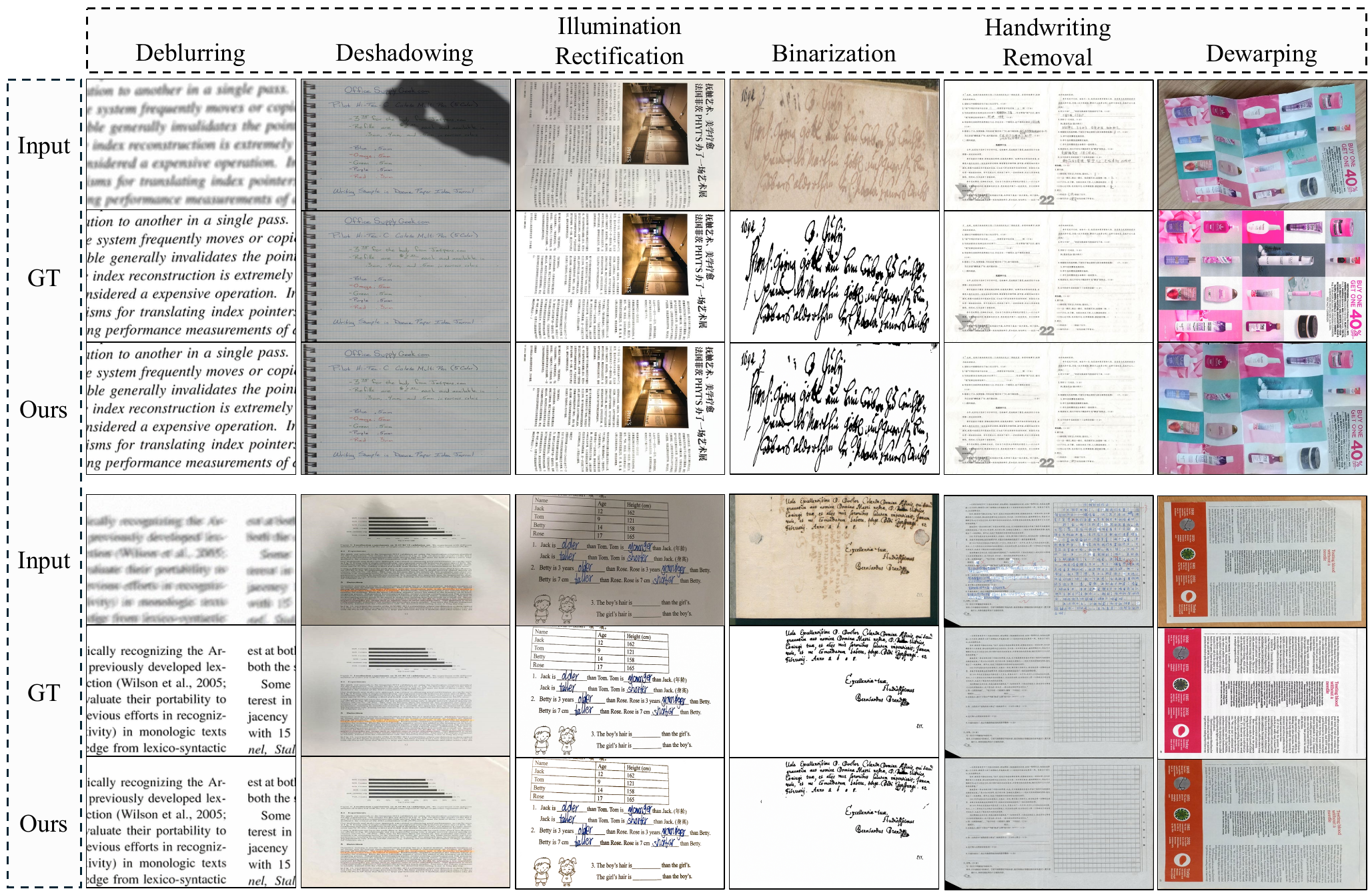}
\caption{The visualization of Uni-DocDiff's results on the six tasks. Each pair of adjacent images represents the original degraded image, GT images and the restored image, respectively.}
\label{visulize}
\end{figure*}

\subsection{Evaluation Metrics}

Deblurring, deshadowing and illumination rectification use PSNR and SSIM as evaluation metrics. To facilitate comparison, similar to previous methods, the dewarping task uses multi-scale structural similarity (MS-SSIM) \cite{you2017multiview}, Local Distortion (LD) \cite{wang2003multiscale}, and Align Distortion (AD) \cite{ma2022learning} as evaluation metrics. Specifically, MS-SSIM is the weighted sum of SSIM values computed over an image pyramid at different resolutions. LD utilizes SIFT flow for dense image registration between two images, measuring the average deformation of each pixel. AD enhances MS-SSIM through unified translation and scaling operations while alleviating LD errors using image gradients as weighting factors. For the binarization task, we adopt PSNR, F-measure (FM), and pseudo F-measure (pFM) for evaluation. For handwriting removal, we continue to use the most commonly employed metrics, MS-SSIM and PSNR.

\subsection{Implementation details}

During the Stage 1 training, we crop the images into 256×256 patches, adopt an AdamW optimizer with an initial learning rate of $10^{-4}$, and a weight decay of $5 \times 10^{-4}$ for 200,000 iterations. For the weights of the loss function, $\beta_1$ is set to 1 and $\beta_2$ to 0.1. For the Stage 2 training, to enable the model to handle images of various sizes during inference, we randomly resize the images to sizes between 256 and 1024 pixels for training and an AdamW optimizer with an initial learning rate of $10^{-4}$ and a weight decay of $5 \times 10^{-4}$ for 100,000 iterations.

\subsection{Result}

\begin{table}[ht]
    \centering
    \setlength{\tabcolsep}{1mm}
    \scriptsize
    \caption{Ablation study about Prior Pool, PFM and frequency-aware Loss. The tasks, listed from top to bottom, are deblurring, deshadowing, illu-rec, dewarping, handwriting removal and binarization.}
    \begin{tabular}{c|c|cccc}
    \toprule
    \multicolumn{1}{c}{Metrics} & \multicolumn{1}{c}{Datasets} & \multicolumn{1}{c}{ w/o Prior Pool } & \multicolumn{1}{c}{ w/o PFM } & \multicolumn{1}{c}{w/o Freq-aware Loss} & \multicolumn{1}{c}{ Uni-DocDiff } \\
    \midrule
    \multirow{2}{*}{{\begin{tabular}{c} SSIM $\uparrow$ \\ PSNR $\uparrow$ \end{tabular}}} & \multirow{2}{*}{TDD} & 0.9502 & 0.9275 & 0.9794 & 0.9824 \\
        & & 23.59 & 21.37 & 27.63 & 28.77 \\
    \midrule
    \multirow{4}{*}{{\begin{tabular}{c} SSIM $\uparrow$ \\ PSNR $\uparrow$ \end{tabular}}} & \multirow{2}{*}{Jung's} & 0.9090 & 0.9038 & 0.9082 & 0.9156 \\
        & & 22.36 & 20.57 & 23.04 & 23.93\\
        & \multirow{2}{*}{Kligler's} & 0.9034 & 0.8907 & 0.9280 & 0.9382 \\
        & & 24.26 & 23.54 & 28.16 & 28.56\\
        \midrule
    \multirow{4}{*}{{\begin{tabular}{c} SSIM $\uparrow$ \\ PSNR $\uparrow$ \end{tabular}}} & \multirow{2}{*}{DocUNet} & 0.7513 & 0.7455 & 0.7661 & 0.7682\\
        & & 16.77 & 16.16 & 17.11 & 18.22\\
        & \multirow{2}{*}{RealDAE} & 0.9273 & 0.9121 & 0.9419 & 0.9485\\
        & & 23.54 & 21.65 & 24.76 & 24.97\\
        \midrule
        \multirow{3}{*}{{\begin{tabular}{c} MSSSIM $\uparrow$ \\ LD $\downarrow$ \\ AD $\downarrow$ \end{tabular}}}
        &\multirow{3}{*}{DIR300} & 0.6331 & 0.6228 & 0.6591 & 0.6573\\
        & & 5.45 & 6.43 & 5.39 & 5.30\\
        & & 0.207 & 0.228 & 0.202 & 0.203\\
        \midrule
        \multirow{2}{*}{{\begin{tabular}{c} SSIM $\uparrow$ \\ PSNR $\uparrow$ \end{tabular}}} & \multirow{2}{*}{EnsExam} & 0.9308 & 0.9299 & 0.9556 & 0.9685\\
        & & 32.46 & 27.50 & 34.32 & 36.23\\
        \midrule
        \multirow{3}{*}{{\begin{tabular}{c} FM $\uparrow$ \\ pFM $\uparrow$ \\ PSNR $\uparrow$ \end{tabular}}}
        &\multirow{3}{*}{DIBCO'18} & 86.31 & 89.39 & 89.80 & 90.32\\
        & & 91.59 & 0.9250 & 0.9314 & 93.84\\
        & & 17.59 & 19.30 & 19.33 & 19.76\\
        \hline

    \end{tabular}
    \label{ablation}
\end{table}

\begin{figure}[tb]
\centering
\includegraphics[width=0.48\textwidth]{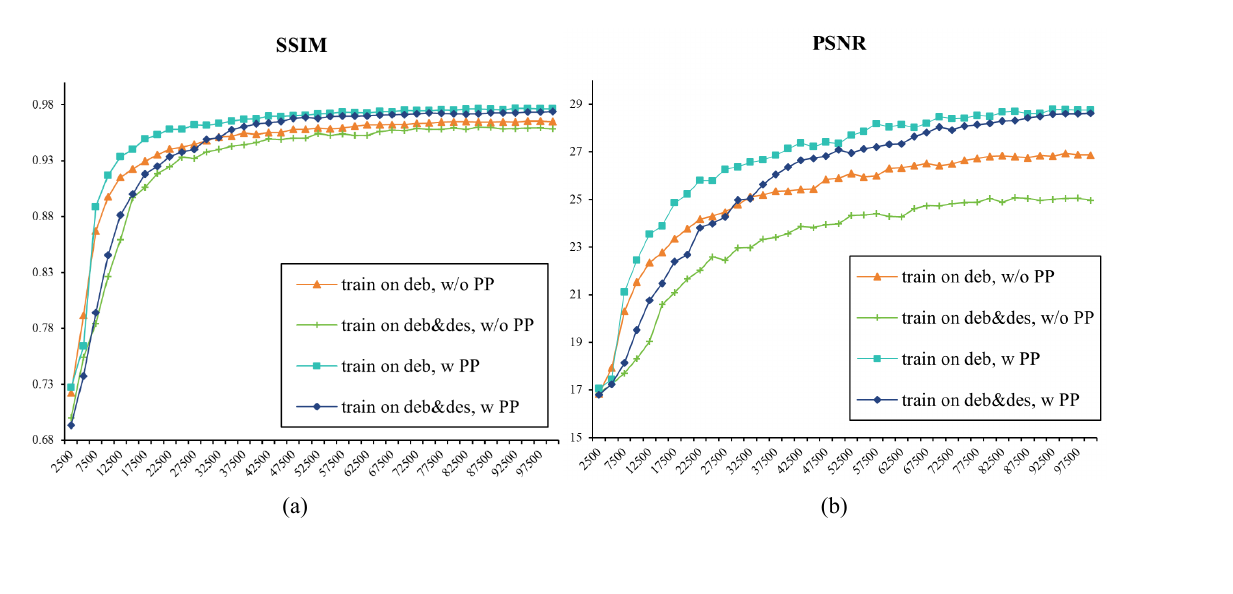}
\caption{Analysis of Inter-Task Interactions During Training. The X-axis represents the number of training steps. We conduct both deblurring task and joint deblurring \& deshadowing tasks, with and without the Prior Pool.}
\label{interface}
\end{figure}

We conduct a comprehensive quantitative comparison across multiple tasks, against both the current state-of-the-art task-specific models and the existing unified restoration model, DocRes, as shown in Tab. \ref{sota_compare}. The results show that our method achieves the best performance on the majority of datasets and metrics, including all metrics on the TDD, DocUNet, RealDAE, DIR300, Jung's Dataset, and EnsExam datasets. In addition, we achieve the second-best results on the SSIM and PSNR for the Kligler dataset, as well as on the FM and PSNR metrics for the DIBCO'18 dataset. This fully demonstrates the effectiveness of our method and its capability to handle multiple tasks.

As shown in Fig. \ref{visulize}, we present the visualization results of the Uni-DocDiff model across six tasks. Uni-DocDiff effectively addresses the given task requirements, successfully removing degradation while preserving the original layout and character shapes, thus delivering high-quality results that meet the desired standards. 
Additionally, our model demonstrates exceptional discriminative capabilities. As shown in Fig. \ref{visulize}, in the illumination rectification task, it effectively preserves handwritten text in document images, while in the handwriting removal task, it precisely eliminates regions containing handwritten text.


\begin{table}[tb]
    \centering
    \caption{Ablation study about task scalability.}
    \begin{tabular}{ccc}
    \toprule
    \multicolumn{1}{c}{Metrics} & \multicolumn{1}{c}{MSSSIM $\uparrow$} & \multicolumn{1}{c}{PSNR $\uparrow$} \\
    \midrule
    DocRes & 0.8886 & 22.79 \\
    Ours & 0.9502 & 32.75 \\
        \hline
    \end{tabular}
    \label{ablation2}
\end{table}

\subsection{Ablation studies}

\paragraph{Prior Pool, PFM and Frequency-aware Loss.} To validate the effectiveness of Prior Pool, PFM, and the frequency-aware loss, we conduct ablation studies for each of these components in Tab. \ref{ablation}. Specifically, (1) for w/o Prior Pool, the prior features are replaced with a set of learnable parameters; (2) for w/o PFM, the module is substituted with convolutional blocks that have the same structure but different parameters for different tasks, of which the feature maps are concatenated in each stage; (3) for w/o Freq-aware Loss, the frequency-aware loss is removed during training. The results highlight the importance of these three designs. Prior Pool and PFM both contribute to the effectiveness of multi-task training by mitigating task interference, while enabling the model to leverage cross-task knowledge transfer. The frequency-aware loss improves model performance across most tasks, with the exception of dewarping where it remains unchanged. To further demonstrate the impact of Prior Pool settings on training, we visualize the training curves for the deblurring task under various settings in Fig. \ref{interface}. Without Prior Pool (w/o PP), combining the deshadowing task with deblurring task in training  degrades the model's performance of deblurring. By leveraging the Prior Pool, the model of joint training achieves equivalent deblurring performance with the model trained solely on deblurring task, and it surpasses the latter when further trained, demonstrating the capability of Prior Pool in tasks interference mitigation.

\paragraph{Task scalability.} In evaluating task scalability, we conduct a comparative analysis between our method and DocRes \cite{zhang2024docres}. Our experimental approach involved initially training models on five document restoration tasks, excluding handwriting removal. Subsequently, we froze all parameters except for the Prompt Feature Module (PFM) and train specifically on handwriting removal task. For the DocRes baseline, we implement a multi-layer convolutional architecture to generate visual prompts for handwriting removal, which are then concatenated with input images and processed by the restoration network. During this phase, we maintain frozen parameters in the restoration network while exclusively training the prompt generator. The quantitative results presented in Tab. \ref{ablation2} demonstrate that our method exhibits superior task scalability compared to DocRes.

\begin{figure}[tb]
\centering
\includegraphics[width=0.45\textwidth]{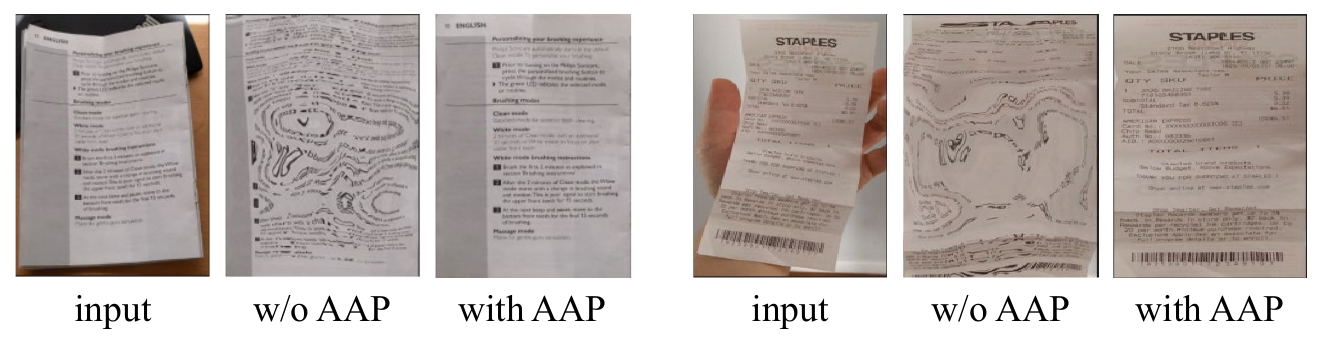}
\caption{Visualization of the dewarping effects with and without AAP.}
\label{aap}
\end{figure}

\paragraph{AAP operation.} For the dewarping task, we observe that the dewarping results will significantly degrade when the training and testing image sizes are inconsistent. To address this issue and enable the model to handle dewarping for arbitrary input sizes, we introduce AAP operator in CPB to fix the size of the feature maps, which effectively solve the problem, as shown in the Fig. \ref{aap}. With AAP, the model can effectively rectify documents of different scales, even if it has not been trained on those specific scales. Without AAP, this capability is not achieved.

\section{Conclusions}

We propose Uni-DocDiff, a unified and highly scalable document restoration model based on diffusion, handling various tasks including dewarping, deblurring, deshadowing, illumination rectification, binarization, and handwriting removal. To mitigate interference between multiple tasks in an all-in-one model, we design a comprehensive Prior Pool and introduce the Prior Fusion Module to adaptively select and integrate relevant priors at specific stages, which makes our model highly scalable across various tasks. Additionally, we develop an efficient dual-stream architecture to decouple coordinate prediction from pixel prediction spaces. Extensive experiments demonstrate that our model achieves satisfactory performance on multiple tasks. In the future, we will focus on considering more task categories and the complete relationships of inter-task interactions.

Although we have conducted experiments to validate our method's effectiveness in addressing task interference, further experimental verification on a broader range of tasks is still needed. Additionally, introducing tasks such as document watermark removal and stamp erasure can enrich the variety of tasks. In the future, we will delve deeper into more tasks and their inter-task interactions.

\begin{acks}
 This work is supported by the National Natural Science Foundation of China (Grant NO 62376266 and 62406318).
\end{acks}

\bibliographystyle{ACM-Reference-Format}
\bibliography{sample-base}

\end{document}